\documentclass[conference]{IEEEtran}
\IEEEoverridecommandlockouts

\usepackage{cite}
\usepackage{amsmath,amssymb,amsfonts}
\usepackage{algorithmic}
\usepackage{caption,subcaption,graphicx}
\usepackage{textcomp}
\usepackage{xcolor}
\usepackage{mathrsfs}
\usepackage{pgfplots}

\usepackage{amsmath}
\usepackage{amssymb}

\usepackage{textcomp}
\usepackage{xcolor}
\usepackage{mathrsfs}
\usepackage{booktabs} 
\usepackage{tikz}
\usepackage{dblfloatfix}
\usepackage{bigstrut}
\usepackage{multirow}
\usepackage{color}
\usepackage{fancyhdr}
\usepackage{listings} 
\usepackage{changepage}
\usepackage{lipsum}

\usepackage{balance}
\usepackage{enumitem}
\usepackage{comment}
\usepackage{breqn}
\usepackage{dblfloatfix}
\usepackage{booktabs} %
\def\BibTeX{{\rm B\kern-.05em{\sc i\kern-.025em b}\kern-.08em
    T\kern-.1667em\lower.7ex\hbox{E}\kern-.125emX}}
\begin{document}

\title{Can GAN Generated Morphs Threaten Face Recognition Systems Equally as Landmark Based Morphs?\\
	\huge - Vulnerability and Detection }
\author{Sushma Venkatesh$^\dag$,  Haoyu Zhang$^\dag$, Raghavendra Ramachandra$^\dag$,  Kiran Raja$^\dag$, Naser Damer$^\ddagger$,  Christoph Busch$^\dag$ \\
$^\dag$Norwegian Biometrics Laboratory, Norwegian University of Science and Technology (NTNU), Norway\\  
	$^\ddagger$Fraunhofer Institute for Computer Graphics Research IGD, Darmstadt, Germany.\\
	\{\tt\small sushma.venkatesh; haoyu.zhang; raghavendra.ramachandra;kiran.raja;christoph.busch\} @ntnu.no\\
	\{\tt\small naser.damer\}@igd.fraunhofer.de\\
}

\maketitle

\begin{figure}[b]
\parbox{\hsize}{\em

}\end{figure}

\begin{abstract}
The primary objective of face morphing is to combine face images of different data subjects (e.g. a malicious actor and an accomplice) to generate a face image that can be equally verified for both contributing data subjects. In this paper, we propose a new framework for generating face morphs using a newer Generative Adversarial Network (GAN) - StyleGAN. In contrast to earlier works, we generate realistic morphs of both high-quality and high resolution of 1024$\times$1024 pixels.  With the newly created morphing dataset of 2500 morphed face images, we pose a critical question in this work. \textit{(i) Can GAN generated morphs threaten Face Recognition Systems (FRS) equally as Landmark based morphs?} Seeking an answer, we benchmark the vulnerability of a Commercial-Off-The-Shelf FRS (COTS) and a deep learning-based FRS (ArcFace). This work also benchmarks the detection approaches for both GAN generated morphs against the landmark based morphs using established Morphing Attack Detection (MAD) schemes. 
\end{abstract}


\section{Introduction}
Due to the widespread deployment of biometric-based identification and verification of individuals, it is essential to observe a biometric characteristic that is reliable, user-friendly and easy to capture. Face biometrics is well suited for this purpose due to its popularity and widespread use for biometric authentication. Moreover we consider the ease of capturing from a distance in a non-intrusive manner and also the recently achieved high recognition accuracy. These properties further enable that face recognition is to be used in various applications that are attributed with high security requirements like border control. However, biometric face recognition systems (FRS) are known to be highly vulnerable to presentation attacks (aka., spoofing attacks) against the capture device \cite{ramachandra2017presentation}. In addition face recognition system can be deceived during the enrolment process by providing manipulated images \cite{ferrara2014magic}.

Among the different types of attacks against FRS, face morphing has gained momentum because of the high impact it poses on border control security. The morphing process enables a malicious actor to generate a morphed image by using an accomplice's face image in a seamless manner \cite{ferrara2014magic}. The process introduces a significant threat to the border control scenario as it is easy to obtain a passport document with a morphed image. This fact is also due to the limitations of current passport issuance protocols in which digital images are submitted in a self-supervised manner by an applicant for passport renewal through web services in countries like New Zealand, Estonia and Ireland. In other countries there exists no live enrolment in the passport renewal process, on the contrary the facial image is provided by an applicant in printed form and is subsequently scanned and re-digitized. This leaves an opportunity for the applicant to morph the face image prior to submitting it in the passport application.

%
\begin{figure}[t!]
	\centering
	\includegraphics[width=1\linewidth]{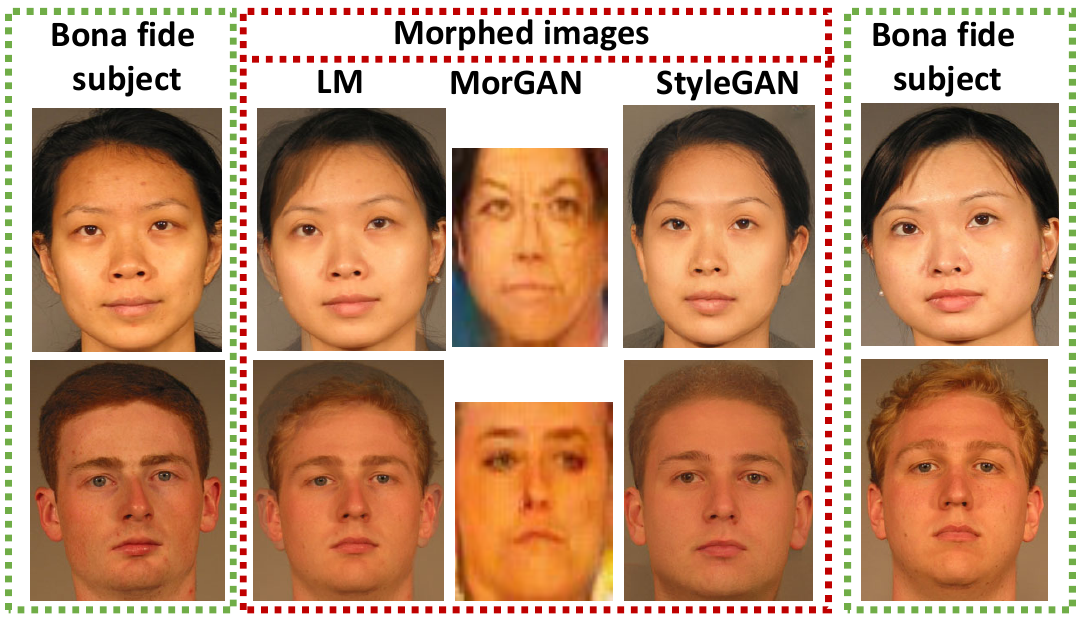}
	\vspace{-3mm}
	\caption{Comparison of morphed images generated using LandMark (LM) , MorGAN and StyleGAN}
	\label{fig:illustration_morph_images}
\end{figure}

\subsection{Morph generation process and limitations}
Early work on face morphing attacks \cite{ferrara2014magic} demonstrated the vulnerability of FRS with respect to morphed facial images, while to the same extend human experts could be fooled \cite{Morph_PLUSONE} \cite{Ferrara2016}.  Following the recent works towards detecting morph attacks on both digital image and re-digitized (print-scanned) image 
\cite{Raghavendra2016, raghavendra2017transferable, raghavendra2017face} we must state that this area of research is still in a premature state. The crucial part of a morphing attack is the generation of high quality morphed facial image, which is ICAO compliant and can attack a deployed FRS with high probability. In the literature, there exist two different ways of generating morphed face images namely (a) Landmark based morphed face generation and (b) Deep learning-based morphed face generation. In landmark-based morph generation, given two images, the landmarks of both facial images are obtained and the Delaunay triangulation is generated for both images. Subsequently alpha blending is performed to obtain a single morphed image based on averaged Delaunay triangles. The majority of the recently published literature is based on open-source morphing tools \cite{raghavendra2017face} which are based on landmark constrained Delaunay triangles.

A deep learning-based approach in contrast involves synthesizing a morphed face image by using a Generative Adversarial Network (GAN).  Limited works are reported in the literature on using GAN for morph generation\cite{MorGAN}. The first reported work in this direction is based on the MorGAN \cite{MorGAN} in which morphed images are generated corresponding to a image resolution of 64$\times$64 pixels. Recently, the morphed images generated using MorGAN were super-resolved to have a incrementally larger dimension of 120$\times$120 pixels \cite{realistic_dreams_MorGAN}. It is important to note that the images generated using both approaches incorporating GAN \cite{MorGAN} \cite{realistic_dreams_MorGAN} are not ICAO compliant and hence have very limited use in real-life attacks.  Irrespective of the morph image generation approach, it is essential that one needs to generate a high-quality image that can pose a high threat potential, when presented to a human expert in the control procedure while the passport issuance is carried out or to surpass a FRS during Automatic Border Control (ABC) crossing scenario.   

\begin{figure}[htbp]
	\centering
	\includegraphics[width=0.8\linewidth]{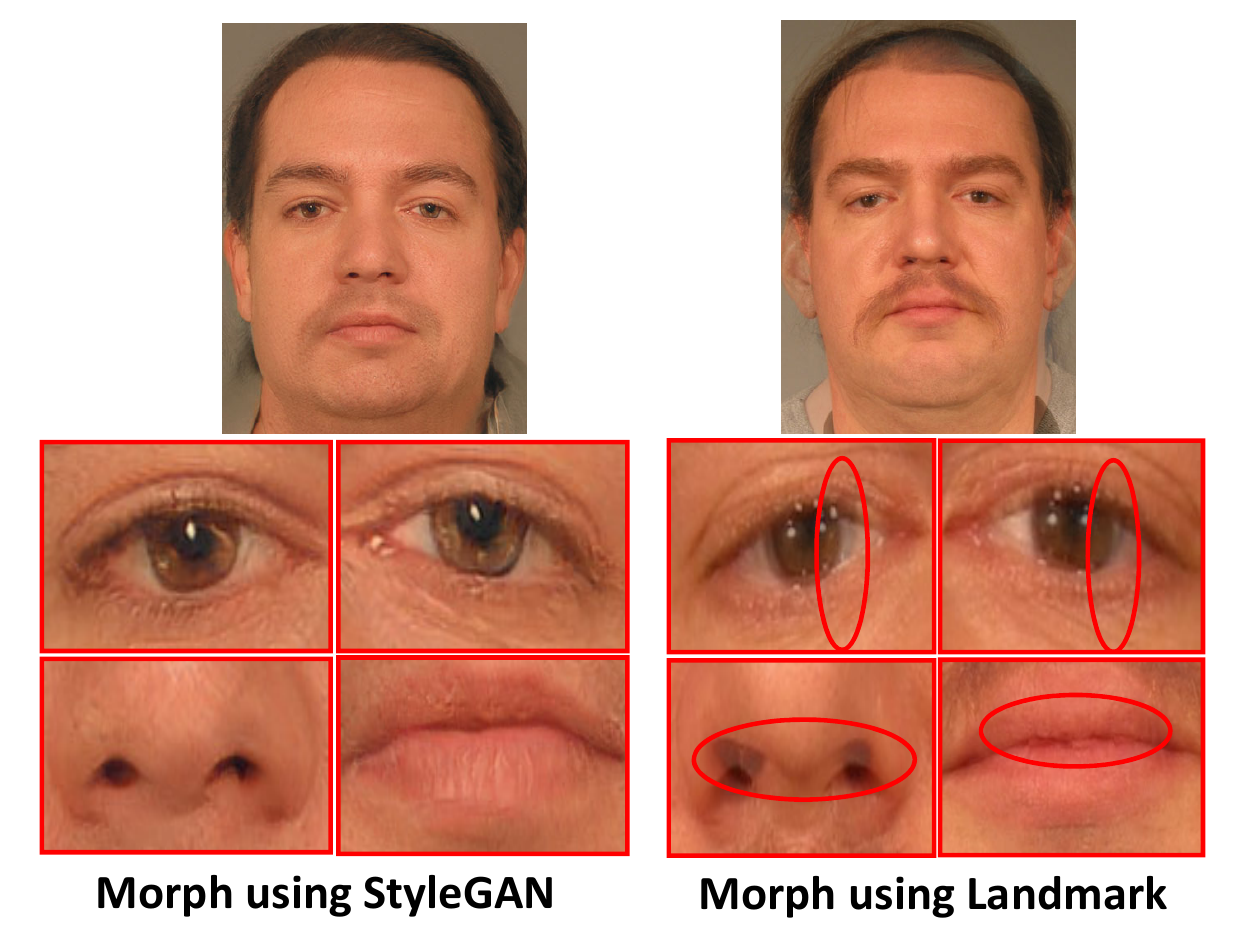}
	\vspace{-3mm}
	\caption{Illustration of minimal artifacts in morphed images generated using StyleGAN versus landmark based face morphing.}
	\label{fig:StyleGAN_LMA_difference_image_5}
\end{figure}

Motivated to address the limitation of low quality images generated by the previous GAN architectures, in this work, we present a new approach to generate high quality morph images. The recent improvement made in GAN architectures has enabled us to generate a high quality facial images with a resolution of 1024$\times$1024 pixels using StyleGAN \cite{karras2019style}. This is achieved by embedding the images into latent space which is further optimized to synthesize the high quality and high resolution image \cite{image2style_GAN}. As illustrated in Figure \ref{fig:illustration_morph_images} the morphed images generated using StyleGAN can be observed to be superior in terms of quality, resolution and visual depiction. 

Further, as noted from Figure~\ref{fig:StyleGAN_LMA_difference_image_5}, a number of artifacts can be easily handled in an automatic manner with the newly proposed approach, which is capable of suppressing visual artifacts. The clear superiority of the newly proposed approach can be noted around the iris regions, where double edges are inherently dealt with. While it well known fact that landmark based morphs threaten FRS to a high degree \cite{raghavendra2017face,ferrara2014magic}, one can easily conclude the amount of extra time and resources that is anticipated to make the morphs visually appealing by removing the artifacts.

While the superior quality of face images can be achieved through the newly proposed approach and eventually reaching compliance to ICAO standards, we raise some fundamental questions. 
\begin{itemize}
	\item Despite the high quality of morphed images do they scale up to threaten a FRS in the same manner as the landmark based morphs, which typically exhibit large artifacts?
	\item To what degree can current MAD mechanisms detect such GAN based attacks on FRS, when the processing is limited to the digital domain? 
\end{itemize}

In the course of answering the above questions, we can summarize the contributions of this work as follows:
\begin{itemize}
	\item A new approach to generate morphed face images using the StyleGAN is presented.
	\item A new face morphing dataset comprising of $2500 \times 3 = 7500$ morphed images is generated using the StyleGAN and MorGAN approach. In order to compare the new approach using the GAN methodology, this work also constructs a corresponding landmark based morph dataset.
	\item To quantify the threats from GAN based morphed face images, a comprehensive vulnerability analysis is conducted using both, a commercial FRS (COTS) and an open-source FRS (ArcFace).
	\item In order to give an insight into the detection challenges of such attacks, this work also reports a detailed evaluation of MAD mechanisms on both GAN based and landmark based morphed face images. 
\end{itemize}

In the rest of the paper, Section \ref{sec:morph_generation} describes the morph generation process proposed in this work using StyleGAN. Section \ref{sec:experiments} provides the details regarding the quantitative experiments indicating the vulnerability of FRS and the detection challenge. With remarks on future works in this direction, we draw the conclusion in Section~\ref{sec:conclusion}.

\section{Morphed Face generation using StyleGAN}
\label{sec:morph_generation}

\begin{figure}[htbp]
	\centering
	\includegraphics[width=1\linewidth]{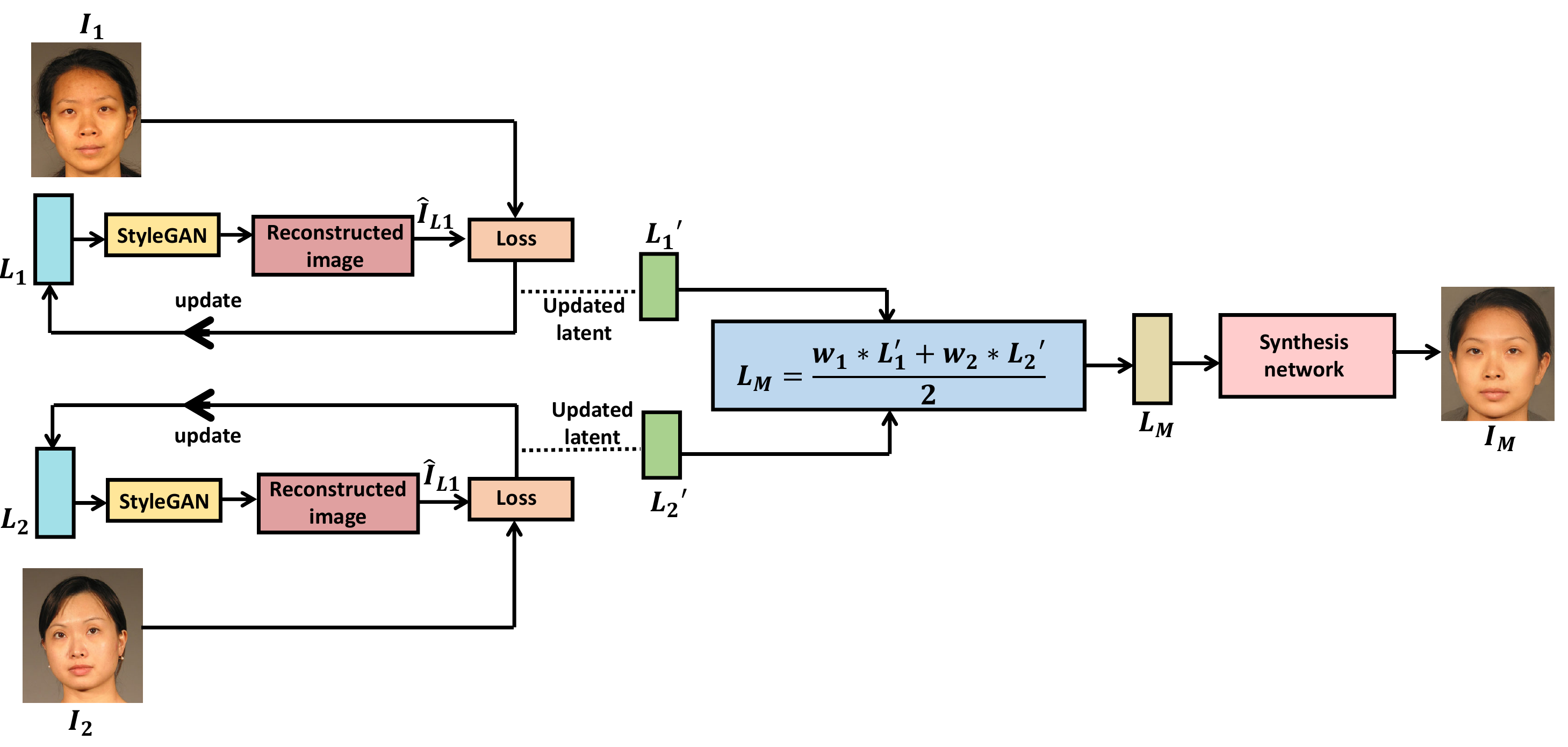}
	\caption{Block diagram of the morphed face image using StyleGAN}
	\label{fig:StyleGAN_proposed_method_1}
\end{figure}

In this section, we present the StyleGAN based face morph generation to achieve high quality face morphs. Figure \ref{fig:StyleGAN_proposed_method_1} depicts the block diagram of the proposed framework for the morphed face  generation using a StyleGAN architecture\cite{karras2019style}. Given the latent code $L_{1}$ of the faces, the StyleGAN \cite{image2style_GAN} maps the inputs to an intermediate latent space ($W$) through the mapping network. The mapping layer consists of $8$ fully connected layers that are serially connected. In this work, we force a strategy to embed the face image into the latent space ($W_{f}$), which is inspired by earlier work \cite{image2style_GAN}. This process enables us to synthesize the data-subject-specific morphed face. The embedded latent space for a particular face is then passed through the synthesis network consisting of $18$ layers, in order to control the adaptive instance normalization (AdaIN).
As a direct result, we obtain the representation in $18$ multiple latent spaces, each with a dimension of 512, which is further concatenated. For a given face image $I_1$, the embedding is carried out by optimizing a loss function that measures the similarity between $I_1$ and the reconstructed image $\hat{I}_{L_{1}}$ using the corresponding latent code $L_{1}$. To maintain the perceptual fidelity a loss is computed as the weighted combination of VGG-16 perceptual loss \cite{image2style_GAN} as given below: 
\begin{equation}
	PL = min\sum_{i = 1}^{4} \frac{\lambda_{i}}{N_{i}} ||F_{j}(\hat{I}_{L_{1}}) – F_{j}(I1)||_{2}^{2}
	\label{eqn:Loss}
\end{equation}

Where, $F_{j}$ is the feature output of VGG-16 layer $conv1_1$, $conv1_2$, $conv3_2$ and $conv4_2$ respectively, ${\lambda_{i}} = 1$ and $N_{j}$ is the number of scalars in the $j^{th}$ layer.  The optimization is carried out using Adam optimizer with a $\beta_{1} = 0.5$. 

We have  selected the perceptual loss based on the visual quality of the morphed image that can reflect the suitability for border control applications. 
Let the final reconstructed image correspond to $I_{1}$ and $L_{1}$ be $\hat{I_{1}}$ and the corresponding updated latent code be $ L_{1}^{'}$.  We follow the same procedure mentioned above for the second image $I_{2}$ to get again a reconstructed image  $\hat{I_{2}}$ and the corresponding updated latent code denoted $L_{2}^{'}$. The morphing operation is carried out by averaging the latent code as follows: 
\begin{equation}
	L_{M} = \frac{w_{1}*L_{1}^{'} + w_{2}*L_{2}^{'}}{2}
	\label{eqn:Mor}
\end{equation}
Finally, $L_{M}$ is passed through the synthesis network to generate the morphed image that has a resolution of $1024 \times 1024$ pixels, where $w_{1}$ and $w_{2}$ indicate the weights, which we have chosen to be $w_{1} = w_{2} = 0.5$.

\subsection{Differences of proposed approach with earlier works}
In contrast to earlier works \cite{MorGAN}, to avoid the bias of morph generation with known set (closed-set), the StyleGAN is trained using the disjoint face dataset from FFHQ dataset\cite{karras2019style} consisting of high quality face images.  As it can be observed from Figure \ref{fig:illustration_morph_images}, the morphed face images generated using StyleGAN have higher perceptual fidelity as compared to MorGAN based morphed images and are equally comparable to landmark based morphed generation. 
It can be noticed that, the MorGAN \cite{MorGAN} based morph generation indicates low-quality images that are not ICAO complaint rendering them not suitable for passport applications.  As a secondary note, the MorGAN based images also indicate a poor visual similarity to the contributing subjects, while landmark based morphs exhibit stronger artifacts that are clearly visible in  Figure \ref{fig:illustration_morph_images}.  

Intrigued by the high fidelity of morphed face images, we take a detailed analysis guided by a sample image to compare it against the landmark based morph generation. As observed in Figure \ref{fig:StyleGAN_LMA_difference_image_5},  the ghosting artifacts in landmark based morphing can be prominently seen due to the misalignment of landmarks leading to several artifacts, especially in the ocular, mouth and nose region. It is interesting to observe that the proposed StyleGAN based morph generation did not create any perceptual noise.The example demonstrates the high quality of the generated image, when compared to the MorGAN based approach. 

While contrary to a landmark based morphed faces, the proposed StyleGAN based morph generation does not indicate a strong geometrical resemblance as it is the case for a landmark based approach. Motivated by such visual observations and superior quality of morph images achieved with the proposed approach and accounting for the lower geometrical resemblance of contributing subjects, we conduct a detailed analysis of threats to FRS as detailed in the next section.

\section{Experiments and results}
\label{sec:experiments}
In order to measure the impact of the proposed approach of morph generation, we first create a new dataset of morphed images created from 140 unique data subjects. With the newly generated morph dataset, we first investigate and report the vulnerability of FRS and compare it with the vulnerability reported in similar earlier work using MorGAN \cite{MorGAN} and traditional landmark based morphing. Further, we also analyze the detection potential of morphed faces generated using the proposed framework with StyleGAN.

\subsection{Database Generation}
We introduce a new morphed face database created from 140 individuals that include 47 female and 93 male data subjects. The facial images are derived from the FRGC-V2 face database \cite{FRGC_DB}. The newly generated database is sub-divided into two sets for training and testing that consists of independent data subjects with no overlap between the splits. The training set consists of 690 bona fide images and 1190 morphed images. The testing set consists of 580 bona fide and 1310 morphed images. To effectively analyze the vulnerability and provide a comparison to earlier works, we have generated morph images using three different techniques, which include (i) Landmark-Based (ii) MorGAN and (iii) proposed StyleGAN approach. Care is exercised to generate morphed images  with similar facial appearance within same gender category. Additionally, to guarantee high quality of the newly generated dataset constraints of high quality illumination and no pose is imposed before creating the morphs. The guidelines laid out in earlier works  \cite{raghavendra2017face} \cite{scherhag2017biometric} are followed to obtain a database of high relevance for morphing attack detection.

\subsection{Evaluation Metrics for Vulnerability Analysis}
We measure the vulnerability of FRS following the guidelines of Frontex and setting the operating threshold to FAR = 0.1 $\%$ (for both FRS). We further follow the realistic evaluation protocol where the morph image is created by using two face images corresponding to a malicious actor and and an accomplice. We compute the vulnerability by enrolling a given morphed face image $M_{I_{1,2}}$ and probing the corresponding contributing subjects $I_{1}$ and $I_{2}$ with an image from a different FRGC-session. We further obtain the comparison scores $S_1$ and $S_2$ for both images $I_{1}$ and $I_{2}$ against the morphed image. The morphed image $M_{I_{1,2}}$  is only considered a threat if and only if the comparison scores $S_1$ and $S_2$ succeed to cross the preset threshold at FAR = 0.1$\%$. If the condition is not met, we simply consider that the morphed image is not a real threat as the comparison scores are not able to successfully verify the morphed image against both contributing subjects making the morphing attack not realistic. We term this new metric as \textsl{Fully Mated Morphed Presentation Match Rate (FMMPMR)} and compute it in general form as: 
\begin{dmath}
	FMMPMR = \frac{1}{P} \sum_{M,P}^{} {(S1_{M}^{P} > \tau) AND (S2_{M}^{P} > \tau) \\ \ldots AND  (Sk_{M}^{P} > \tau)}
	\label{Eqa:FMMPMR}
\end{dmath}

Where $P = {1, 2, \ldots, p}$ represent the number of attempts made by presenting all the probe images from the contributing subject against $M^{th}$ morphed image,  $K = {1, 2, \ldots, k}$ represents the number of contributing data subjects to the  constitution of the generated morphed image (in our case  $K=2$), $Sk_{M}^{P}$ represents the comparison score of the $K^{th}$ contributing subject obtained with $P^{th}$ attempt (in our case the $P^{th}$ probe image from the dataset) corresponding to $M^{th}$ morph image and $\tau$ represents the threshold value corresponding to FAR = 0.1$\%$. 

When compared to the existing metric MMPMR \cite{scherhag2017biometric}, the FMMPMR considers the number of attempts (that are assessed jointly with contributing subjects) with regards to each face morphed images and thus reflect the realistic vulnerability of a FRS. The MMPMR \cite{scherhag2017biometric} is designed to measure vulnerability only on the morphed image in a joint set rather than a number of attempts on each morphed image. Hence, the MMPMR fails to reflect the number of attempts (by contributing subjects) made against the corresponding morphing image to determine the vulnerability of FRS.

In this work, the COTS threshold is set at $\tau = 0.5$ based on the NIST FRVT test reports as recommended by the COTS provider  while ArcFace FRS threshold is set at $\tau = 0.36$ base on the face recognition trials on FRGC-v2 dataset. The higher the value of the FMMPMR the higher the threat from morphed images and correspondingly a higher vulnerability of FRS towards morphed images must be stated. 

\begin{figure}[htbp]
	\centering
	\includegraphics[width=1\linewidth]{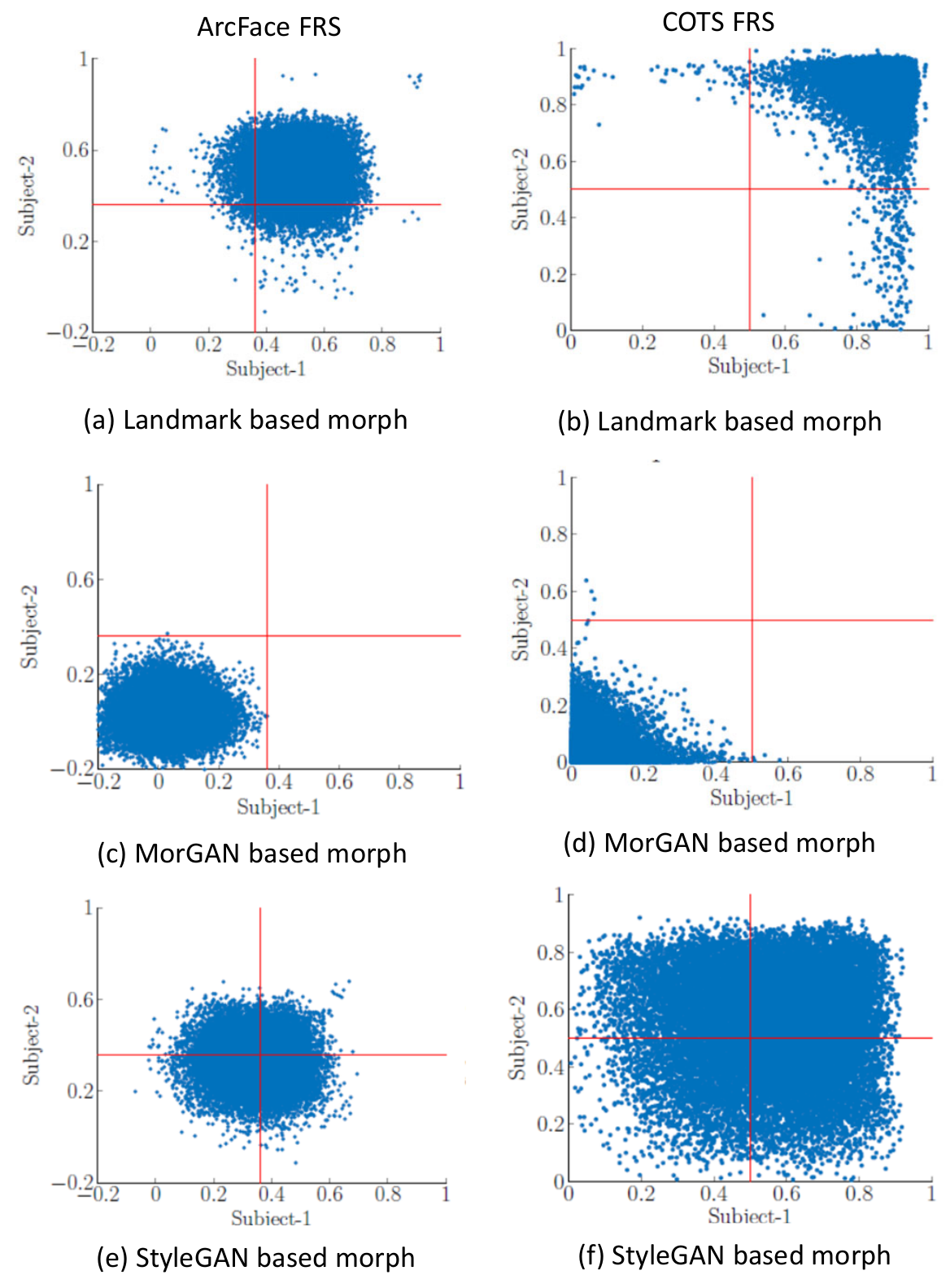}
	\caption{Vulnerability analysis using COTS and ArcFace. The scatter plots represents the comparison scores of morphed face image against two contributing subjects.The red lines  indicate the threshold corresponding to FAR = 0.1\%}
	\label{fig:StyleGAN_plots_Arcface_COTS}
\end{figure}

\subsection{Results from Vulnerability analysis}
In this section, we present the vulnerability analysis using two different Face Recognition Systems (FRS) (i) a Commercial off the Shelf face recognition system (COTS), Cognitec FaceVACS-SDK Version 9.4.2 \footnote{outcome not necessarily constitutes the best the algorithm can do} and (ii) an  Open-source deep learning based FRS (ArcFace).  To effectively benchmark the results we also compare two different State-Of-The-Art (SOTA) morph generation techniques such as landmark based morph generation \cite{UBO_Morphing_Tool} and MorGAN based morph generation \cite{MorGAN}.

Figure \ref{fig:StyleGAN_plots_Arcface_COTS} shows the scatter plot of the comparison scores obtained from two different FRS on images obtained using three different types of morphed face generation approaches. Table \ref{tab:FMMPMR} indicates the quantitative values of both MMPMR and FMMPMR computed from two different FRS for all three cases of face morph generation techniques. Based on the obtained results the key observations made are listed below:

\begin{table}[htbp]
	\centering
	\caption{Vulnerability analysis FMMPMR($\%$) and MMPMR($\%$) }
	\resizebox{1\linewidth}{!}{
		\begin{tabular}{ccccc}
			\hline
			\multirow{2}[4]{*}{Morphing Type} & \multicolumn{2}{c}{FMMPMR (\%)} & \multicolumn{2}{c}{MMPMR (\%)} \bigstrut\\
			\cline{2-5}    \multicolumn{1}{r}{} & \multicolumn{1}{c}{ArcFace} & \multicolumn{1}{c}{COTS} & \multicolumn{1}{c}{ArcFace} & \multicolumn{1}{c}{COTS} \bigstrut\\
			\hline
			Landmark based & 86.04   & 98.91   & 95.76   & 100 \\
			Morph\cite{UBO_Morphing_Tool} & & & & \bigstrut\\
			\hline
			MorGAN\cite{MorGAN}  & 0       & 0       & 0       & 0 \bigstrut\\
			\hline
			StyleGAN & 21.1    & 41.49   & 39      & 64.68 \bigstrut\\
			\hline
		\end{tabular}%
	}
	\label{tab:FMMPMR}%
\end{table}%

The following  are the main observations from our experiments.
\begin{itemize}
	\item Landmark based face morph generation indicates a high threat to FRS (analogously high vulnerability of FRS to such images) compared to that of two other morph generation methods. This can be attributed to the fact that the landmark based morph generation preserves both texture and geometrical structure of the morphed image corresponding to it's  contributing subjects. 
	\item The analysis of the experimental results also show that MorGAN based morph generated images do not pose a severe threat to FRS. The potential reason for this can be due to low quality generated morph image (64 $\times$ 64 pixels). A careful observation of the images also revealed the degradation of texture and geometry in the generated morphed images. As a caveat, we note that the MorGAN network is not re-trained (or fine-tuned) on closed dataset of contributory subjects. The conscious choice was made to investigate the generalisation of GANs for morph face image generation and study the threats. 
	\item StyleGAN based morph generation method shows relatively higher degree of threats when compared with MorGAN. Despite higher threats, the images from proposed approach of morph generation did not compete against the landmark based methods. An introspection into this indicates the quality difference of FFHQ dataset versus the employed FRGC-V2 dataset. Specifically, the pre-trained morph generator is trained on  FFHQ dataset which has very different characteristics than FRGC-V2 dataset leading the network to mimic the characteristics of the FFHQ dataset. Another aspect for the lower degree of threat is due to lack of geometric correspondence of facial structure in morphed faces when compared to that of the landmark based face morphing. The lower geometrical correspondence despite the high visual quality fails in verification stage from FRS.
	\item When compared to ArcFace, the COTS indicates a higher vulnerability for both landmarks and StyleGAN based morph attack detection due to the high accuracy of verifying the subjects in COTS under different data capture conditions as expected in operational scenario. Thus, COTS while making itself robust about certain degree of degraded data, also accepts the morphs to a higher degree.   
	\item  Table \ref{tab:FMMPMR} also indicates the distinction between FMMPMR and MMPMR metric used to quantify the vulnerability. The MMPMR reports high values in comparison to FMMPMR as it does not account for the number of attempts per morphing image. Further, we have also measured the Relative Morph Match Rate (RMMR) \cite{scherhag2017biometric} that can account for the True Acceptance Rate of the FRS. Since both FRS employed in this paper have reached TAR = $100\%$, the RMMR is the same as the FMMPMR/MMPMR. 
\end{itemize}

\begin{table}[htbp]
	\centering
	\caption{Quantitative performance of state-of-the-art MAD techniques on StyleGAN dataset}
	\resizebox{1\linewidth}{!}{
		\begin{tabular}{ccccc}
			\hline
			\multicolumn{1}{c}{\multirow{2}[4]{*}{Morphing Type}} & \multirow{2}[4]{*}{Algorithms} & \multicolumn{1}{c}{\multirow{2}[4]{*}{D-EER(\%)}} & \multicolumn{2}{c}{BPCER(\%) @ APCER } \bigstrut\\
			\cline{4-5}            & \multicolumn{1}{c}{} &         & \multicolumn{1}{c}{=5(\%)} & \multicolumn{1}{c}{=10(\%)} \bigstrut\\
			\hline
			\hline
			\multicolumn{1}{c}{\multirow{4}[8]{*}{}} & HoG-SVM & 10.29   & 17.66   & 10 \bigstrut\\
			\cline{2-5}  Landmark          & LBP_SVM & 15.42   & 29.15   & 22.98 \bigstrut\\
			\cline{2-5}  based Morph \cite{UBO_Morphing_Tool}          & Color Textures & 1.57    & 0.51    & 0.17 \bigstrut\\
			\cline{2-5}            & CAN     & 4.8     & 4.63    & 2.4 \bigstrut\\
			\hline
			\hline
			\multicolumn{1}{c}{\multirow{4}[8]{*}{MorGAN \cite{MorGAN}}} & HoG-SVM & 0       & 0       & 0 \bigstrut\\
			\cline{2-5}            & LBP_SVM & 0       & 0       & 0 \bigstrut\\
			\cline{2-5}            & Color Textures & 0       & 0       & 0 \bigstrut\\
			\cline{2-5}            & CAN     & 0       & 0       & 0 \bigstrut\\
			\hline
			\hline
			\multicolumn{1}{c}{\multirow{4}[7]{*}{StyleGAN}} & HoG-SVM & 0.04    & 0       & 0 \bigstrut\\
			\cline{2-5}            & LBP_SVM & 0.68    & 0       & 0 \bigstrut\\
			\cline{2-5}            & Color Textures & 0       & 0       & 0 \bigstrut\\
			\cline{2-5}            & CAN     & 0.36    & 0       & 0 \bigstrut[t]\\
			\hline
		\end{tabular}%
	}
	\label{tab:Experiment}%
\end{table}%

\subsection{Performance Metrics for MAD}
The performance of Morphing Attack Detection (MAD) techniques are presented using the ISO/IEC 30107-3 metrics \cite{ISO-IEC-30107-3-PAD-metrics-170227} such as Attack Presentation Classification Error Rate (APCER ($\%$)) which defines the proportion of attack images incorrectly classified as bona fide images and Bona fide Presentation Classification Error Rate (BPCER ($\%$)) in which bona fide images incorrectly classified as attack images \cite{ISO-IEC-30107-3-PAD-metrics-170227} along with the Detection Equal Error Rate (D-EER ($\%$)).

\subsection{MAD Detection Performance}
In this section, we report the detection performance of MAD techniques to understand the impact of different types of morphing  techniques. We have therefore selected four different MAD techniques - LBP-SVM \cite{Raghavendra2016}, HoG-SVM \cite{Sushma_WACV2019}, color denoising \cite{Sushma_IPTA2019}, Context aggregation Network (CAN) \cite{Sushma_WACV2019} based on the recent benchmarks. Table \ref{tab:Experiment} indicates the MAD performance on all three different morph generation techniques.

Compared to three different morph generation methods, landmark based technique indicates a relatively high challenge for the detection techniques, when compared to that of GAN based techniques. However on the same kind of morph generation approach, the recent technique based on color texture indicates the lowest error rates with D-EER(\%) of 1.57(\%). While it is noted that the GAN generated morphs are easier to detect, a possible reason can be attributed to the residual noise \cite{karras2019analyzing} that is associated with GAN in generating these morphed images. Even though StyleGAN can generate a high quality images with a resolution of 1024$\times$1024 pixels, the inherent noise in the generated morph images make enables to detect them. This is not the case for landmark based morph images, which do not contain such characteristic noise.

\subsection{Limitations and Future Directions}
Observing the results from the empirical evaluation of different approaches of morph generation both for threats to FRS and ability to detect the morphs, we note certain limitations in the current work as listed below. 
\begin{itemize}
	\item The GAN based morph generation does not impose the landmark correspondence leading to high quality images but not with high facial similarity in geometrical appearance to contributing subjects. This has lead to lower threat to FRS in comparison to landmark based morphs. Future works in this direction can focus on imposing such a constraint in the latent space, in order to increase the threat to FRS.
	\item Despite the accuracy of MAD being very high, it can be primarily attributed to digital pixel level information helping to detect the attacks. A print and scan of the the same morphed images can further reveal the real challenge in detecting the morphing attacks as the print-scan cycle looses the pixel level soft-information in the image.
\end{itemize}
The future works in this direction will lead to establishing the real threat landscape on FRS from the GAN generated morphed face images.

\section{Conclusion}
\label{sec:conclusion}
This work investigated the feasibility of generating high quality morph generation and proposed a new approach using StyleGAN. The proposed approach resulted in morphed face images with a dimension of 1024$\times$1024 pixels and no visual artifacts. To indicate the real threat potential to FRS, the morphed face images generated from proposed StyleGAN were analyzed using a commercial FRS and an open-source FRS. Further, to provide a fair comparison to earlier works, MorGAN and Landmark based approaches were benchmarked on the same set of data by creating a new morphed face database. The set of experiments clearly indicate the that StyleGAN based morphed face images do show threats to FRS but to a much lower degree as compared to traditional landmark based morph generation techniques. While detecting the attacks stemming from GAN approaches is relatively easy in the digital domain, the real challenge of detecting them after the print-scan process is still not explored. In summary, we answer the question - \textit{Can GAN Generated Morphs Threaten Face Recognition Equally as Landmark Based Morphs?}, our experimental results indicates with a clear no in digital domain alone.

{\small
	\bibliographystyle{ieee}
	\bibliography{sushma-IWBF-2020-StyleGAN-200216}
}

\end{document}